# Handwritten Urdu Character Recognition using 1-Dimensional BLSTM Classifier

Saad Bin Ahmed*, Saeeda Naz**, Salahuddin Swati***, Muhammad Imran Razzak *

*King Saud Bin Abdul Aziz University for Health Sciences, Riyadh, Saudi Arabia.
**Hazara University, Department of Information Technology, Mansehra, Pakistan.
**Higher Education Department, KPK, Pakistan.
***COMSATS Institute of Information Technology, Abbottabad, Pakistan.

Email: (ahmedsa, razzaki)@ksau-hs.edu.sa, saeedanaz292@gmail.com, salahuddin@ciit.net.pk

**Abstract:** *The recognition of cursive script is regarded as a subtle task in optical character recognition due to its varied representation. Every cursive script has different nature and associated challenges. As Urdu is one of cursive language that is derived from Arabic script, that's why it nearly shares the same challenges and difficulties even more harder. We can categorized Urdu and Arabic language on basis of its script they use. Urdu is mostly written in Nasta'liq style whereas, Arabic follows Naskh style of writing. This paper presents new and comprehensive Urdu handwritten offline database name Urdu-Nastaliq Handwritten Dataset (UNHD). Currently, there is no standard and comprehensive Urdu handwritten dataset available publicly for researchers. The acquired dataset covers commonly used ligatures that were written by 500 writers with their natural handwriting on A4 size paper. We performed experiments using recurrent neural networks and reported a significant accuracy for handwritten Urdu character recognition.*

**Keywords:** R*ecurrent neural networks, optical character recognition, cursive offline handwriting.*

---

## 1. Introduction

The term document image analysis is a process that interprets digital documents. The digital documents may comprise textual data processing (OCR, layout analysis) and graphical processing (line, region and shape processing). Furthermore, it can be in form of synthetic data or in scanned form. The specialized technology named Optical Character Recognition (OCR) is used to read characters of rendered data either taken from scanned text or in printed form. There exists two sorts of style in most of the language i.e., cursive and non-cursive. In non-cursive style the recognition of character is not be a problem but when there is cursive text, it is an evident challenge to separate characters and make them recognized by classifier. Moreover, the dataset for most of cursive scripts cannot be available publicly for evaluation purpose.

The OCR is a significant area in document image analysis. It reads document images and translates them into searchable text. The OCR systems are capable of recognizing characters and thereby words and sentences from document image. The researchers have shown great interest since past decade to address the potential problems that exist in this particular field [1–8]. Dataset is considered as collection of document images particulary in image analysis field. To propose dataset is a primary step for development of reliable OCR in the process of document image analysis. The role of dataset is crucial to evaluate the performance of state-of-the-art techniques. There exists various Latin datasets like IAMoffline [15], MNIST [3], UNLV-ISRI [12], for Latin word and character recognition. The Arabic script based languages like Arabic, Urdu, Persian, Sindhi, Pashtu, Balochi, Uigher and Jawi are used by the considerable world population. Arabic script based languages character recognition are still considered as challenging task in field of OCR due to the complexity of this script i.e. cursive writing, graphospasm multiplication, joiner and non-joiner property of letters, overlap in ligature and with other ligature etc. Moreover, there are various writing styles for Arabic script based languages but the most common writing styles are Naskh (for Arabic, Sindhi, Pashto) and Nasta'liq ( for Urdu and Persian and Pakistani Punjabi).

Nasta'liq writing style is more difficult than Naskh due to its complexity (like diagonality, filled loops, false loop, no fixed baseline and large variation of words/subwords) over Naskh writing style [16–18]. The work done for Arabic cannot be applied directly on Nastaliq writing style. There is no significant OCR system specialized for Urdu language has been reported to date. However, there has been an increase in interest of the research community especially from the Indian subcontinent to address Urdu Nastaliq OCR problem.

This paper is organized in various sections. Section 2 summarized detail about dataset for Arabic script based languages followed by indication of different challenges that exist in Urdu language is depicted in Section 3. These challenges envision research ideas especially in Urdu language. Section 4 presents details about acquisition of proposed Urdu Nastal'iq dataset and the preproccesing that has applied on acquired data. Section 5 elaborates about the scenario where we can evaluate proposed dataset. The dataset evaluation followed by discussion on performed experiments has been detailed in Section 6. The result depicted in Section 7 while Section 8 concludes our effort and recommendation for future work has been proposed.

## 2. Related Work

The dataset for Arabic script based languages [2,19,20] are limited and not been addressed the problems of cursive script in detail. Some efforts have been reported for Arabic character recognition [19,21,22,23]. There are different datasets available for Arabic scripts, summarised in Table 1 but unfortunately, there is no publicly available handwritten dataset for Nastaliq to research community. While, character set is almost same for both scripts (i.e., Naskh and Nasta'liq) but we cannot use Nasta'liq as a replacement for Naskh due to complexity involved in prior script. Efforts are being made to standardize the dataset of Urdu language for the purpose of comparing different available state-of-the-art techniques. One such effort is made by Centre for Pattern Recognition and Machine Intelligence (CEPARMI) [24] to develop a handwritten database from different sources that consists of isolated digits and forty four isolated characters, numeral strings and fifty seven words (related to finance field), five special symbols, and dates in Urdu Nastaliq. Other efforts are being reported by Image understanding and Pattern Recognition Group at the Technical University of Kaiserslautern, Germany to generate synthetic data of Urdu language, whose contents were taken from leading Urdu newspaper of Pakistan named Jang [25]. The Jang newspaper prints the text in Alvi Nasta'liq script which covers the political, social and religious issues.

Another very useful dataset is prepared by [37]. They reported cursive Urdu character recognition results by Bidirectional Long Short Term Memory (BLSTM) networks. They performed experiments on synthetic data gathered from one of leading newspaper from Pakistan. They applied window based approach for taking feature value and provide them to classifier. They reported better accuracies with respect to position and without position information of Urdu character.

Essoukri et al. developed an Arabic relational database for Arabic OCR systems named ARABASE [4] and evaluated their two systems by using their proposed database. One system was evaluated printed Arabic writing using the Generalized Hough Transform while the other system was evaluated with handwritten Arabic script using Planar Hidden Markov Model. The candidates were supposed to provide their name as author. This information was also used in preparation of database for off-line Arabic handwriting (AHDB) [26] which is open source database. They collected samples from 100 writers. This database contains words used in writing legal amounts in banks. It is considers as the most popular Arabic words. Another database named CENPARMI is freely available dataset that consist of 3,000 handwritten cheque images [27].
It consists of labelled 29,498 sub-word images, 15,175 digit images, and 2,499 legal amount images. The database is designed to facilitate automatic cheque reading research for the banking and finance sectors.

**Table 1** Summary of Arabic, Persian and Urdu Databases.

| Database | Type | Size | Availability |
| --- | --- | --- | --- |
| ARABASE [4] | Arabic handwritten | Not reported | Aavailable |
| AHDB [26] | Arabic handwritten | Not reported | Not available |
| IFN/ENIT [28] | Arabic handwritten | 26,459 city name | Free available |
| CEDAR [29] | Arabic handwritten | 100 pages, each consist of 150-200 words | Not available |
| CENPARMI [30] | Arabic handwritten | 29,498 subwords, 15,175 digits, a 2,499 digits | Available |
| ERIM [31] | Arabic typewritten and printed text | 750 pages | Not available |
| APTI [32] | Arabic printed text | 45,313,600 word | Available |
| IFN/Farsi [33] | Persian handwritten | 7,271 words | Available |
| FHT [34] | Persian handwritten | 1,000 forms | Availble |
| CLE [35] | Urdu printed ligatrues | 15000 subwords | Available |
| UPTI [36] | Urdu synthetic text | 10,000 subwords text lines not reported | Avalible |

## 3. Challenges in Urdu Language

The Urdu language is written from right to left like Arabic and Persian as represented in Fig. 1. Every character can occur in isolation, at initial, middle or at final position in a word. The joining of letters represents cursive nature of such languages. The characters may join or not join with its preceding or/and subsequent letter, due to joiner and non-joiner property of cursive text. The joiner character may appear at initial, middle, isolated or at final position as represented in Fig. 2. These characters joined with preceding and subsequent letters in a word when it occurs in the middle position. As an initial letter in a word, it joined its subsequent letter. When appears as a final letter of word, it joins with its preceding letter. When joiner letters occur at initial position or at middle position it may completely change its shape as depicted in Fig. 3. The non-joiner character appears in its full shape as isolated or at final position in a single word. Urdu word comprised of ligatures that make a word meaningful. Associate ligatures of same word pose huge challenge for researchers to address. For such complicated scripts, context is important to learn. The numeric characters in Urdu is written from left to right but Urdu writing starts from right to left as represented in Fig. 4. In the Naskh font, there are only four shapes but in case of Nasta'liq, we are facing more shape-context than Naskh. The common issues related to Urdu language are mostly the same as its ancestor languages have e.g., context sensitivity, characters overlap, word segmentation and text line segmentation. But these challenges require more considerable efforts to address issues in Nasta'liq script as compared to Naskh.

The diagonally, multiple baseline, placement of dots, number of dots association, kerning, stretching, false and filled loop are the unique challenges [17] for detail of challenges in Nastaliq) in Urdu language and need to be tackled down. Fig. 4 depicts normal format of Urdu characters and words. There is scarcity of database in Nastaliq script, the UNHD offline handwritten database provides strength to the problem of Urdu OCR particularly in Nastaliq script.

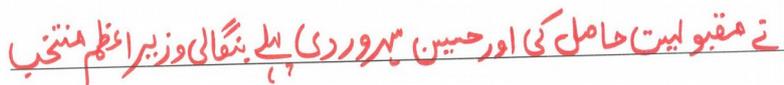

**Fig. 1** Urdu handwritten sentence in Nasta'liq font.

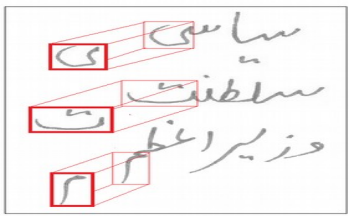

**Fig. 2** Urdu words with last character appeared in actual shape.

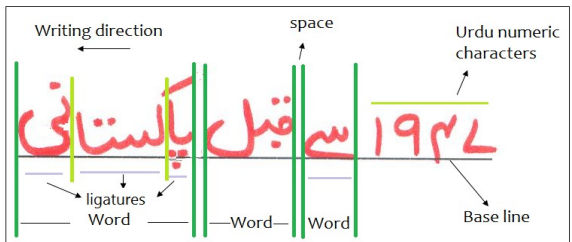

**Fig. 3** Urdu characters with its isolated, initial, middle and final shapes that may occur in a Urdu word [37].

**Fig. 4** Representation of Urdu text with numeric characters, words and ligatures

## 4. UNHD Database

For the purpose to preform Urdu handwritten OCR development and evaluation, we present comprehensive handwritten Urdu dataset named UNHD for Urdu Nastaliq Handwritten Dataset. UNHD database covers all Urdu characters and ligatures with different variations. In addition, the Urdu numeric data is also taken from authors. This dataset consist of ligatures which comprised up to five characters. The taken dataset can applies as handwritten character recognition as well as writer identification. The initial data was taken in red colour to ensure integrity of taken samples. The integrity of data refers to maintain its pixel value during noise removal. The noise is usually in black colour and it is easy to determine it if text has other than black color. The vocabulary of UNHD database spans digits, numbers, and part of speech noun, verb, pronoun etc., but in Urdu Nasta'liq font. Later, in the enhanced version we acquired samples from school and college students. Moreover, we also collected data from office going individuals to ensure that we must have all variability of handwrittten samples from all individuals associated to any field and from every age. The data collected from 500 writers (both male and female).

In this way we have broaden our collected text from 48 unique text lines to 700 unique textlines including Urdu numerals and Urdu constraint handwritten samples. we have more than 6000 Urdu handwritten textlines.The text lines have been written with three variations like, on the baseline, without baseline and slanted on the page. Each individual was trained before taking sample and were asked to write provided text in a natural way. Thus, each individual wrote 48 text lines. Table 2 shows complete depiction of gathered data for UNHD. We had given 6 blank pages to each individual with their identification and page numbers were mentioned. Moreover, to mitigate the processing and to correct slant and skew, we also provide some pages with marked baselines to writer as shown in Fig. 5. Each individual is asked to write the provided printed text. The UNHD dataset text is obtained with identification of each individual that will helps in writer identification. The dataset consist of 3,12000 words written by 500 candidates with total of 10,000 lines. There are approximately 624 words written by a single author. We have tried to cover Urdu text as maximum with character variability according to it's position and writing style. There are approximately 1,87200 number of characters exist in UNHD dataset to date.

**Table 2** UNHD dataset details

| UNHD details | Statistics |
|---|---|
| Total no. of writers | 500 |
| No. of text lines per page | 8 or 5 text lines |
| Total no. of text lines | 10000 |
| Number of words written by single writer | approx. 624 |
| Total no. of words | 312000 |
| Total no. of characters | approx. 187200 (Consider 6 characters per word) |

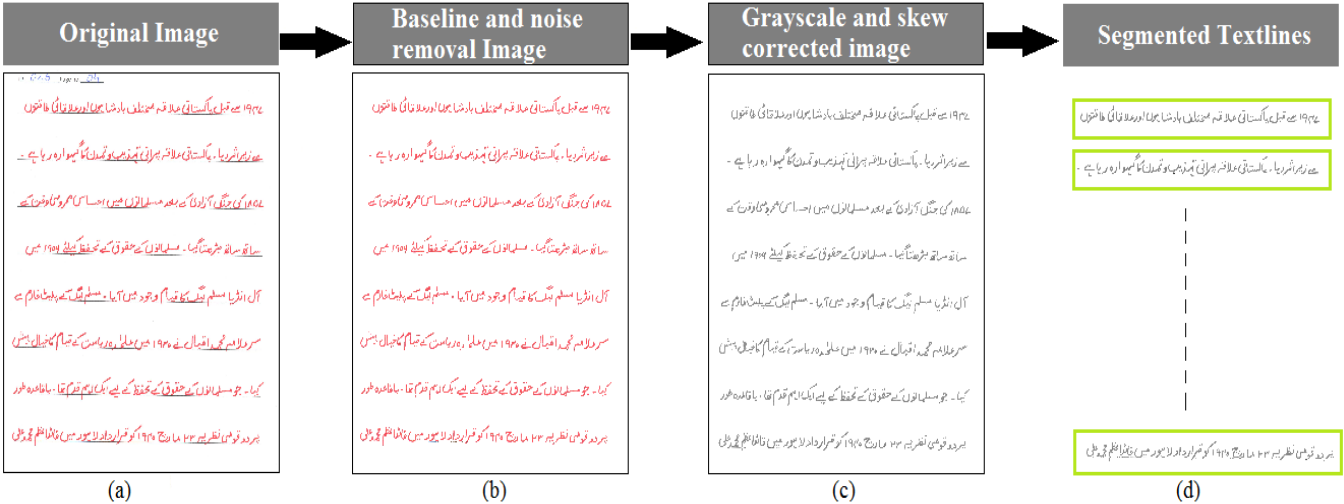

**Fig. 5** Pre-processing of Urdu text document.(a) represents actual data acquired using scanner having identification number and page number. (b) shows removal of noise and baseline (c) is a gray scaled and unscewed image which eventually passed to the program of text line segmentation. (d) segmented text line.

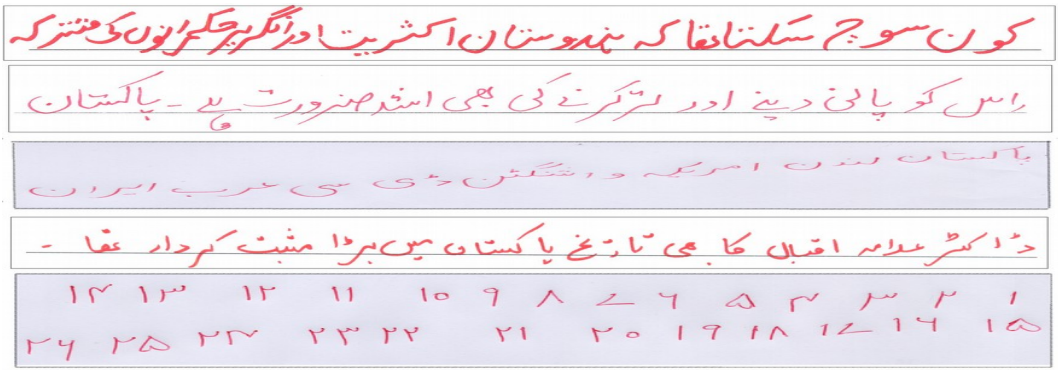

**Fig. 6** Different samples of input images with noise and without noise.

**4.1 Image Acquisition and Pre-processing**

The following pre-processing steps applied on UNHD database that includes removal of baseline (if there is baseline in the document) and noise, grey scale conversion, skew detection and correction, text lines segmentation (shown in Fig. 5) and database labelling/annotation.

The image acquisition is the first step in the workflow of character recognition to acquire image using some hardware based sources like scanners, cameras, tablets etc. The pages scanned with a HP Scanjet 200 flatbed scanners using 300 dpi (ideal for subsequent steps of OCR processing). Each line is extracted from given page automatically and given a label with 9 letters including digits and dash "-" symbol e.g., ddd-dd-dd (d=0, 1, 2 . . .9). According to this coding, the first three digits represents writer's id (e.g., 001. . . 300), next two digits show page id (e.g., 01. . . 06) and last two digits tell about the serial number (e.g., 01. . . 08) for different text lines in a given page. This coding scheme can be coped with the future growth of database and also helpful in sorting and searching the contents of database with efficiency.

### 4.1.1 Noise and Skew Removal

The scanned pages are used for removal of baseline using the colour information. The median filter is applied for suppression of noise while maintaining the sharp edges. As in Urdu, it is cumbersome to learn each shape of every character in presence of variations with respect to one character. Therefore, we dealt with pixel values to accommodate different shapes information. The original image was converted into grey scale. The detection and correction of skew is a crucial part for segmentation step in OCR. In the literature, different methods reported [5, 14, 18, 40]. In our dataset, we got data on drawn baseline on the page from 100 writers for lessen skew in text line and 200 writers wrote without baseline. For skew correction we used horizontal projection method [14] for all data from 500 writers. In this method the image is project at different angles and calculates the variance of horizontal projection. Horizontal projection is the sum of each row of the image. The horizontal projection of un-skewed image will likely have the maximum value. The stepwise detail of whole process is depicted in Algorithm 1.

*Algorithm 1*

```
INPUT: Gi (Gray scale skewed image)
OUTPUT: Gi (Gray scale skew corrected image)
    Begin
        Bi=:Binary(Gi)
        hp=:Horizotalprojection(Bi)
        max=:var(hp)
        angle=:0
    For theta=:-value to value
        temp_Bi=:rotate(Bi,theta);
        hp=:Horizotalprojection(Bi)
        temp=var(hp);
        IF max<temp
            max=temp;
            angle=theta;
        End
    End
        Gi=: rotate(Gi,theta);
    End
```

### 4.1.2 Text-line Segmentation

The text lines are segmented by projection profile in horizontal direction. At first, we applied projection profile method to get the positions of each pixel. After taking text pixel position we segment lines from grey scale image. The resulted text lines with their ground truth used as an input to the classifier for learning the character patterns. The Algorithm 2 shows the pseudocode of textline segmentation.

*Algorithm 2*

```
INPUT: Gi (Gray scale image)
OUTPUT: SLi (Segmented lines)
 BEGIN
    Bi=:Binary(Gi)
    hp=:Horizotalprojection(Bi)
    j=:1
    lw=:0
    pt=:0
    ln=:1
    While j<=length(hp)
        IF hp(j)>0
        pt=:j-1;
        While hp(j)>0
```

```
                    lw=:lw+1;
                    j=:j+1;
            End
            SLi(ln)=:crop(Gi(pt:pt+lw,:))
            ln=:ln+1
        End
        j=:j+1
        End
End
```

### *4.1.3 Dataset Labelling*

In pattern recognition and machine learning, the ground truth information uses with actual image in supervised learning that provides standard to learn patterns in the image. It is considered as backbone for supervised learning tasks. It is an essential part of database for OCR's experiment in segmentation step. To evaluate the performance of recognition system, ground truth needs to be labelled correctly. Each text line is described with the help of ground truth text file. The *utf8* encoding of every non-Latin character has been declared. Like other non-Latin languages, Urdu also has its *utf8* encoding characters regardless to its position. In reading Urdu, character positions are vital in determination of word for reader. As specified in Fig. 3 suppose the character meem occurred at different positions. There is no need to declare different utf8 codes with respect to each position of meem. So, we have only one code of meem for its every position, graphically it may have different shapes but its *utf8* code would be same. We labelled every character with its four possibilities of occurrences in Urdu word. With reference to Fig. 3, the character appeared in isolation is labelled as *meem_iso*, at initial position as *meem_i*, at middle position *meem_m* and at final position as *meem_f* respectively.

## 5. Dataset Research Scenarios

Based on UNHD offline dataset there are some typical scenarios where document analysis tasks can be performed. Our recommendations are mentioned as follows.

1. The dataset is stored and tagged with identification of user, thus can be used to perform writer identification on data sample. For, user identification can be performed dataset form 500 individual.
2. It is a cumbersome task to gather data from writers. The UNHD offline database will be used and apply different techniques for segmentation of text-lines and words into sub-word/ligatures.
3. To evaluate the potential of state of the art techniques on cursive scripts like Urdu, the proposed dataset can be used.
4. Another variation of UNHD dataset usage is to take geometrical information of every character and maintain the information in separate table against every character. The captured information in this way can be trained and used for character recognition.
5. To determine ligatures from Urdu words is another research area which can be performed using UNHD database.

## 6. Dataset Evaluation

The acquired dataset is completely applicable on any type of classifier. The classifier we used is precisely defined below.

### 6.1    Recurrent Neural Network

The classifier we proposed for our recognition system is Recurrent Neural Networks (RNNs). As learned from literature [6, 40] the basic constraint of Multi-Layer Perceptrons (MLPs) is to map input into output vector without considering the previous computations at output unit while RNN has flexibility of tracing back previous computations. In this way history also takes a part in computations at hidden layer. The internal state of the network is retained by looped connection that makes an influence at output level. The RNNs are meant to retain the previous sequence information [6]. Fig. 7 shows complete depiction of the proposed system.

The RNNs are meant to use their feedback connections that exist in hidden layer for the purpose to retain most recent calculation that contributes in weight calculation of current node in a sequence. When given input is complex and large then the time lag for retaining computation would be difficult to maintain and we can lose information as a result. In this particular situation that information which requires for longer period of time would be vanished from cell's memory. To address this problem the concept of Long Short Term Memory (LSTM) networks introduced [41]. In LSTM, the hidden layer memory cells were replaced by memory blocks with additive multiplicative units. These multiplicative units are responsible for maintaining the information for longer period of time and vanish the gradient information when it is no longer required by the sequence. In some situation, we need to predict the future point in a sequence. To consider the future point in time we applied LSTM in forward and backward direction for the purpose to have the context of a text at given point in a sequence. We applied Bidirectional Long Short Term evaluation of proposed dataset. The Section 6 shows the strength of a given technique.

### 6.2    Text page Segmentation

The given text page was segmented into text lines by projection profile method. The features of an image play a crucial role in text-line recognition. The image is segmented into candidate regions and features are extracted from each candidate region as marked in Fig. 8.

After applying pre-processing on UNHD handwritten dataset, we performed segmentation of text lines. The window size of 30 * 1 (x-height) traverses over the given text line image to get corresponding pixel values as feature values which is given to classifier for learning as shown in Fig. 7. To test the applicability of UNHD handwritten dataset, we took 7200 text line images as train-set and 4800 text lines were used to validate the training 2400 text-lines were in test-set. The 6.04 to 7.93 percent error was reported on UNHD offline dataset.. The produced results are highly motivated. We performed experiments on the text lines written by 500 authors to assess the potential of RNNs on handwritten cursive data as mentioned in Table 3.

**Table 3:** Dataset contribution w.r.t. writers

| Number of writers contributed | Text lines | Dataset distribution |
|---|---|---|
| 300 | 6400 | Train set |
| 120 | 1760 | Validation set |
| 80 | 1840 | Test set |

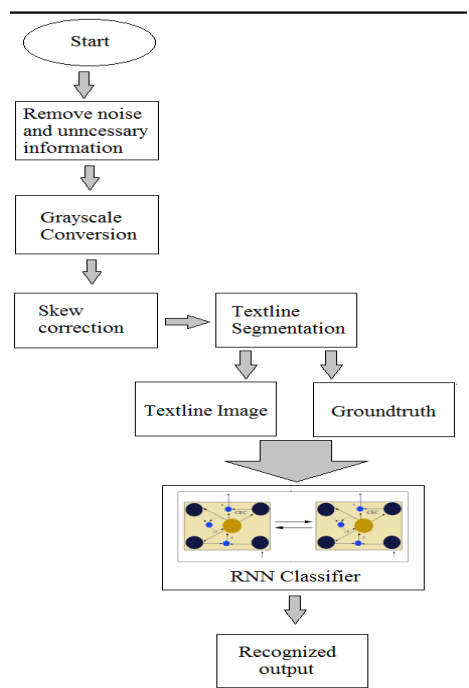

**Fig. 7** Proposed recognition system

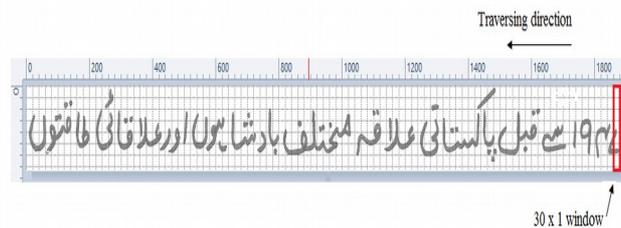

**Fig. 8** Normalized to x-height 30 x 1 window size in red colour.

## 7. Result and Discussion

This section provides detail of BLSTM evaluation performed on UNHD offline handwritten dataset name Urdu-Nasta'liq handwritten dataset. The 1-dimensional Bidirectional Long Short Term Memory (BLSTM) networks is used, which is a connectionist classification approach based on RNN [40], LSTM architecture [42] and Bidirectional Recurrent Neural Network (BRNN) [43]. As mentioned before that BLSTM is a RNN approach which is considered ideal for sequence learning. The sequence is crucial to get exact output as desired. There is a need to use such classifier that can address the past sequences for the intention to predict the output symbol at current point in time. Due to constraint of managing context, BLSTM is considered as an attractive choice for learning the sequential data that requires feedback from current or previous temporal sequences.

The dataset is divided into train set, validation set and test set with a ratio as 50% for training, 30% used for validation and remaining 20% was used for test set. The number of authors that took part to make train set, validation set and test set are mentioned in Table 3. The learning curve formed during training of transcription is represented in Fig. 8. As we learned from the curve that there is approximately 5% difference between the training and validation. The difference is gradually increases after 52th epoch. In order to avoid over training we terminate the learning on 60th epoch. The learning is depicted in Fig. 9.

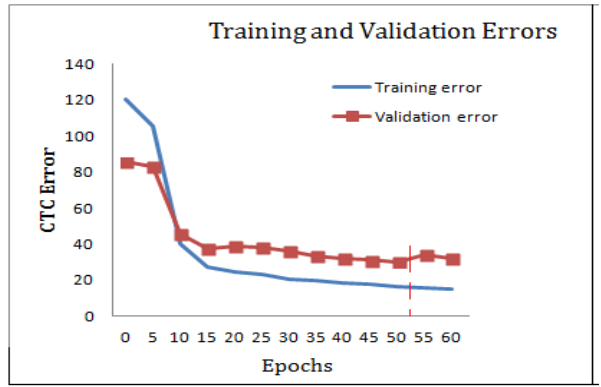

**Fig. 9** CTC error rate during training of transcriptions (at character level). The horizontal red line shows the optimal point after that the distance between the training and validation tends to increase.

We trained network on different BLSTM hidden layer size as shown in Fig. 10. As learned from [5] that size of hidden layer make an impact on learning. We get best performance of RNN by increasing the size of hidden layer neurons i.e., 100. The recognition error tends to increase after 100th hidden layer size. This means that we get optimal result on 100th hidden neurons. However, the time to train network on different hidden memory units will increase by increasing number of units. This indicates that the number of hidden memory units is directly proportional to time.

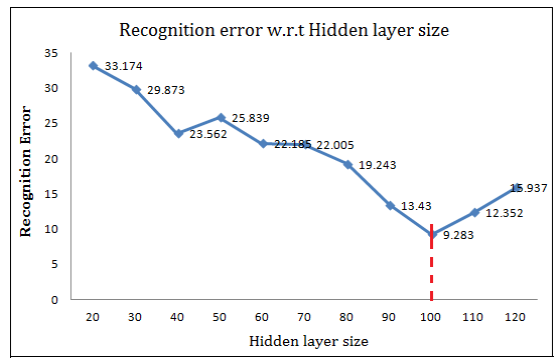

**Fig. 10** The recognition error rate measured on different hidden layer sizes. The dotted red line indicates the best recognition.

## 8. Conclusion and Future Work

In this paper, we proposed new dataset for handwritten Urdu language named Urdu-Nastaliq handwritten dataset. Being a cursive nature, Urdu Nasta'liq font has no standard dataset available publicly. The basic motive of preparing UNHD offline database is to compile Urdu text and make it available to research community free of cost. The data has been gathered from 500 individuals, but it will be extended up to 1000 individuals. Currently, UNHD database covers commonly used ligatures with different variations in addition to Urdu numeric data. Although we achieved very good character accuracy i.e., approximately 6.04 to 7.93 percent error rate. The performance of classifier can be evaluated on 2-dimensional BLSTM. Other future tasks may include writer identification, apply different feature extraction approaches, and apply different classifiers to recognize the text and word recognition with the help of dictionary and language modelling.